\documentclass[runningheads]{llncs}

\usepackage{eccv}
\usepackage{bbm}
\usepackage{eccvabbrv}
\usepackage{graphicx}
\usepackage{booktabs}
\usepackage{wrapfig}
\usepackage[accsupp]{axessibility}
\usepackage{color}
\usepackage{hyperref}
\usepackage{orcidlink}
\usepackage{float}
\usepackage{multirow}

\begin{document}

\title{End-to-End Rate-Distortion Optimized 3D Gaussian Representation} 

\titlerunning{RDO-Gaussian}

\author{Henan Wang\inst{1} \and
Hanxin Zhu\inst{1} \and
Tianyu He\inst{2} \and
Runsen Feng\inst{1} \and
Jiajun Deng\inst{3} \and
Jiang Bian\inst{2} \and
Zhibo Chen\inst{1}}

\authorrunning{H.~Wang et al.}

\institute{University of Science and Technology of China \and
Microsoft Research Asia \and
The University of Adelaide}

\maketitle

\vspace{-2mm}

\begin{abstract}
3D Gaussian Splatting (3DGS) has become an emerging technique with remarkable potential in 3D representation and image rendering. However, the substantial storage overhead of 3DGS significantly impedes its practical applications. In this work, we formulate the compact 3D Gaussian learning as an end-to-end Rate-Distortion Optimization (RDO) problem and propose RDO-Gaussian that can achieve flexible and continuous rate control. RDO-Gaussian addresses two main issues that exist in current schemes: 1) Different from prior endeavors that minimize the rate under the fixed distortion, we introduce dynamic pruning and entropy-constrained vector quantization (ECVQ) that optimize the rate and distortion at the same time. 2) Previous works treat the colors of each Gaussian equally, while we model the colors of different regions and materials with learnable numbers of parameters. We verify our method on both real and synthetic scenes, showcasing that RDO-Gaussian greatly reduces the size of 3D Gaussian over 40$\times$, and surpasses existing methods in rate-distortion performance. 
Our source code is available at \href{https://github.com/USTC-IMCL/RDO-Gaussian}{https://github.com/USTC-IMCL/RDO-Gaussian}.

\end{abstract}

\section{Introduction}
\label{sec:intro}
Recently, 3D Gaussian Splatting (3DGS) has gained increasing popularity owing to its potent capability in novel view synthesis~\cite{kerbl20233d}, establishing itself as one of the most promising 3D representations. Different from neural radiance fields which implicitly model scenes via neural networks~\cite{mildenhall2021nerf}, 3DGS opts for anisotropic ellipsoids to explicitly represent scene properties, where impressive real-time rendering can be achieved. Various follow-ups employ 3DGS in multiple aspects such as anti-aliasing~\cite{yu2023mipsplatting}, dynamic scene reconstruction~\cite{wu20234dgaussians,duan20244d}, and 3D generation~\cite{tang2023dreamgaussian,yi2023gaussiandreamer,liu2023humangaussian}, etc. However, though remarkable results can be obtained, a common drawback of these 3DGS-based methods is their huge storage cost, which hinders their practical applications to a large extent and motivates the study of a compact 3DGS representation.

To achieve a compact representation, pioneer works target to minimize the \textit{rate} of 3DGS while maintaining the \textit{distortion} of rendered results as much as possible~\cite{lee2023compact,fan2023lightgaussian,navaneet2023compact3d,morgenstern2023compact}. For example, several works propose to reduce the number of Gaussians and represent colors as hash grids~\cite{lee2023compact} or distilled parameters~\cite{fan2023lightgaussian}. CompGS~\cite{navaneet2023compact3d} introduces vector quantization based on the K-means to quantize the Gaussian attributes. Although significant rate reduction is achieved, limitations still exist: 1) they minimize the rate under the fixed distortion, however, it often requires rate-distortion trade-offs in the context of data compression~\cite{balle2016end,balle2017end}. 2) these works treat the colors for each Gaussian equally, such as using volumetric grids~\cite{lee2023compact} or Spherical Harmonics (SHs) with the same degree~\cite{fan2023lightgaussian}. This assumption, however, is not always efficient in the cases of different materials, illuminations, etc. For example, it is redundant for diffused objects while necessary for objects that have a view-dependent appearance. 3) the compact representation is obtained by multi-stage optimization~\cite{fan2023lightgaussian}, or off-the-shelf compression tools~\cite{morgenstern2023compact}, which restrict the end-to-end learning from the data distribution.

To mitigate the above limitations, we present end-to-end Rate-Distortion Optimized 3D Gaussian representation (RDO-Gaussian) with flexible and continuous rate control. To this end, we formulate the 3D Gaussian representation learning as a joint optimization of \textit{rate} (R) and \textit{distortion} (D), i.e., R+$\lambda$D, where $\lambda$ governs the trade-off between them. Our scheme is built upon the 3DGS framework~\cite{kerbl20233d} with three proposed components: Gaussian pruning, adaptive SHs pruning, and entropy-constrained vector quantization (ECVQ). Specifically, motivated by that the original training of 3DGS brings numerous redundant Gaussians~\cite{lee2023compact}, in Gaussian pruning, we learn to remove the redundant Gaussians by employing a learned mask for each Gaussian. For the attributes of each Gaussian, we observe that the majority of Gaussian parameters come from the SHs ($76.3\%$), which is not necessary for every Gaussian to have the same degree in the case of different materials and illuminations. Therefore, we introduce adaptive SHs pruning to each Gaussian, allowing each Gaussian to have different degrees of SHs. Following this, to further obtain a compact representation, we convert the covariance and color attributes into the discrete representation with ECVQ, which conducts vector quantization and also allows rate-distortion joint optimization. As a result, the quantized and entropy-encoded parameters form the final 3D Gaussian representation.

We verify our method in multiple benchmarks including both real scenes and synthetic scenes. Results show that our method can achieve similar rendering quality while greatly reducing the Gaussian size compared to existing compression methods. 

The main contributions of this paper can be summarized as:
\begin{itemize}
    \item We formulate the 3D Gaussian representation learning as a joint optimization of \textit{rate} and \textit{distortion} to achieve flexible and continuous rate control.
    \item To realize rate-distortion optimization, we control the rates by introducing dynamic pruning and entropy-constrained vector quantization (ECVQ) that optimize the rate and distortion at the same time.
    \item We for the first time achieve volumetric bit allocation by introducing adaptive SHs pruning.
    \item On both real and synthetic scenes, our method surpasses existing methods in RD performance. The compression ratio can be up to over 40$\times$ compared to the original 3DGS.
\end{itemize}

\section{Related Works}
\label{sec:related}

\subsection{Compact Neural Radiance Fields}
Neural Radiance Fields (NeRF)~\cite{mildenhall2021nerf} has shown remarkable performance for continuous scene representation thanks to its utilization of neural networks, where photorealistic novel views can be generated in a per-scene optimization manner.
However, due to its slow rendering, various methods introduce explicit representations such as points~\cite{peng2021shape,xu2022point}, trees~\cite{yu2021plenoctrees,wang2022fourier} or feature grids~\cite{sun2022direct,fridovich2022plenoxels,muller2022instant} to speed up the rendering process. However, this brings additional storage costs. To reduce the storage costs, some works search for more compact representations~\cite{chen2022tensorf,tang2022compressible,reiser2023merf,rho2023masked}, while others use data compression techniques such as pruning or quantization to compress the feature grids~\cite{takikawa2022variable,girish2023shacira,zhao2023tinynerf,li2023compressing,deng2023compressing}. However, NeRF series with high rendering quality~\cite{barron2022mip,barron2021mipnerf,verbin2022refnerf,zhu2024cmc,zhu2024vanilla} still leverage neural networks as the implicit representation, which is slow in both training and rendering. Such a problem motivates the innovation for novel 3D representation like 3D Gaussian Splatting (3DGS)~\cite{kerbl20233d}.

\subsection{Compact 3D Gaussian}

3D Gaussian Splatting (3DGS)~\cite{kerbl20233d} has gained increasing popularity due to its high visual quality and fast rendering~\cite{kerbl20233d}. However, due to the massive size of the 3D Gaussians, various compression strategies have been developed so far. CompGS~\cite{navaneet2023compact3d} uses K-means vector quantization (VQ) to compress Gaussian attributes. K-means cluster centroids and indexes are periodically updated along with the training of Gaussians. Four codebooks of different sizes are used for vector Gaussian attributes including scales, rotations, base colors and SHs. LightGaussian~\cite{fan2023lightgaussian} adopts a multi-stage compression strategy, including importance pruning, SH distillation, and vectree compression. Specifically, every Gaussian is given an importance score based on its volume, opacity, and hit counts in rendering. Gaussians with low importance scores are pruned and the rest are finetuned to recover performance. After pruning, SHs are distilled from $3$ degrees to $2$ degrees by knowledge distillation to reduce the number of SH parameters. Finally, VQ is applied to attributes of the least important Gaussians, excluding positions and opacities. Positions are compressed by octree compression, as outlined in point cloud compression methodologies~\cite{graziosi2020overview}. The remaining attributes are saved in \texttt{float16} format to reduce storage. Compact-3DGS~\cite{lee2023compact} uses Gaussian pruning, residual VQ, and hash grid-based color representations to compress Gaussians. For Gaussian pruning, each Gaussian is allocated with a learned mask which marks whether the Gaussian is pruned or not. Covariance parameters, including scales and rotations, are compressed by residual VQ. Inspired by grid-based 3D representations like InstantNGP~\cite{muller2022instant}, the authors adopt hash grids to represent colors rather than SHs.

Different from the above methods that obtain the compact representation by only minimizing the rate under the fixed distortion, we formulate the compact 3D Gaussian representation learning as a joint optimization of rate and distortion to achieve flexible and continuous rate control. We also realize volumetric bit allocation by introducing adaptive SHs pruning for the first time.

\section{Methods}
\label{sec:methods}

We build our method on 3D Gaussian Splatting (3DGS)~\cite{kerbl20233d}. In this section, we first give an introduction to the background of 3DGS in Sec.~\ref{sec:3dgs} and then elaborate on the proposed RDO-Gaussian in Sec.~\ref{sec:rdo_gaussian} and \ref{sec:encoding}.

\begin{figure}[tb]
  \centering
  \includegraphics[width=1.0\textwidth]{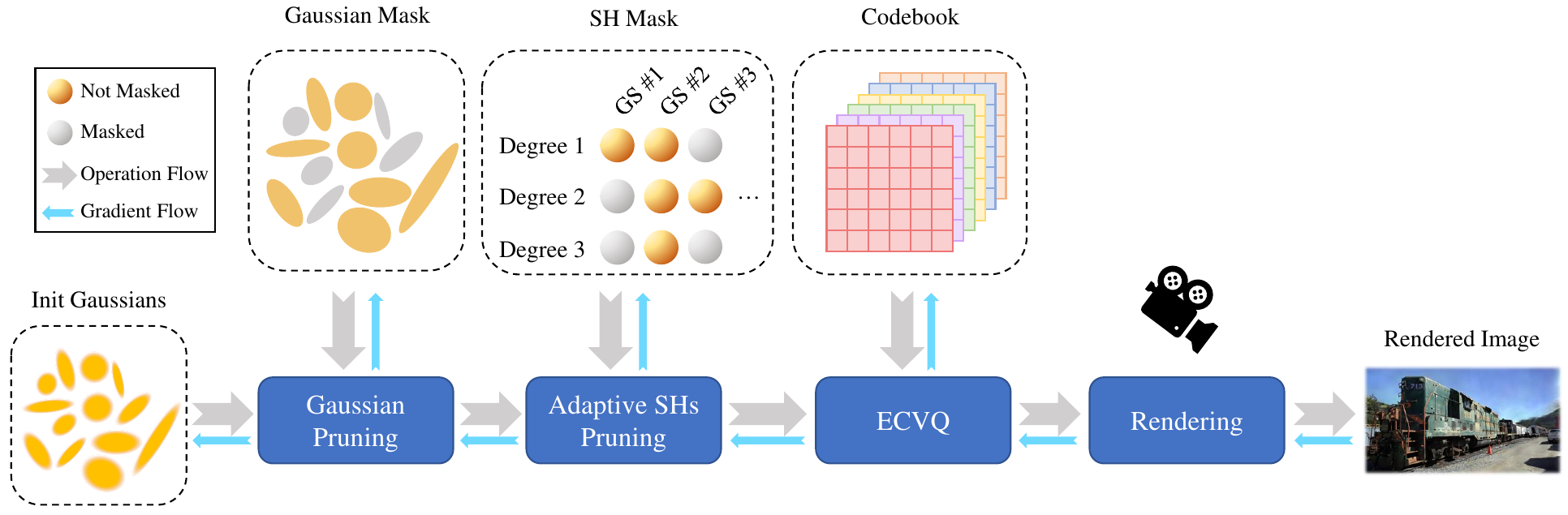}
  \caption{Training pipeline of RDO-Gaussian. We prune both Gaussians and SHs with learnable masks, which eliminates the redundant Gaussians and achieves volumetric bit allocation by adapting different SH degrees to different Gaussians. The pruned Gaussians are quantized by ECVQ to get discrete representations. Gaussian masks, SH masks and codebooks are jointly optimized along with Gaussians.}
  \label{fig:pipeline}
\end{figure}

\subsection{Preliminary: 3D Gaussian Splatting}
\label{sec:3dgs}

\begin{wrapfigure}{r}{0.4\textwidth}
  \centering
  \includegraphics[width=0.4\textwidth]{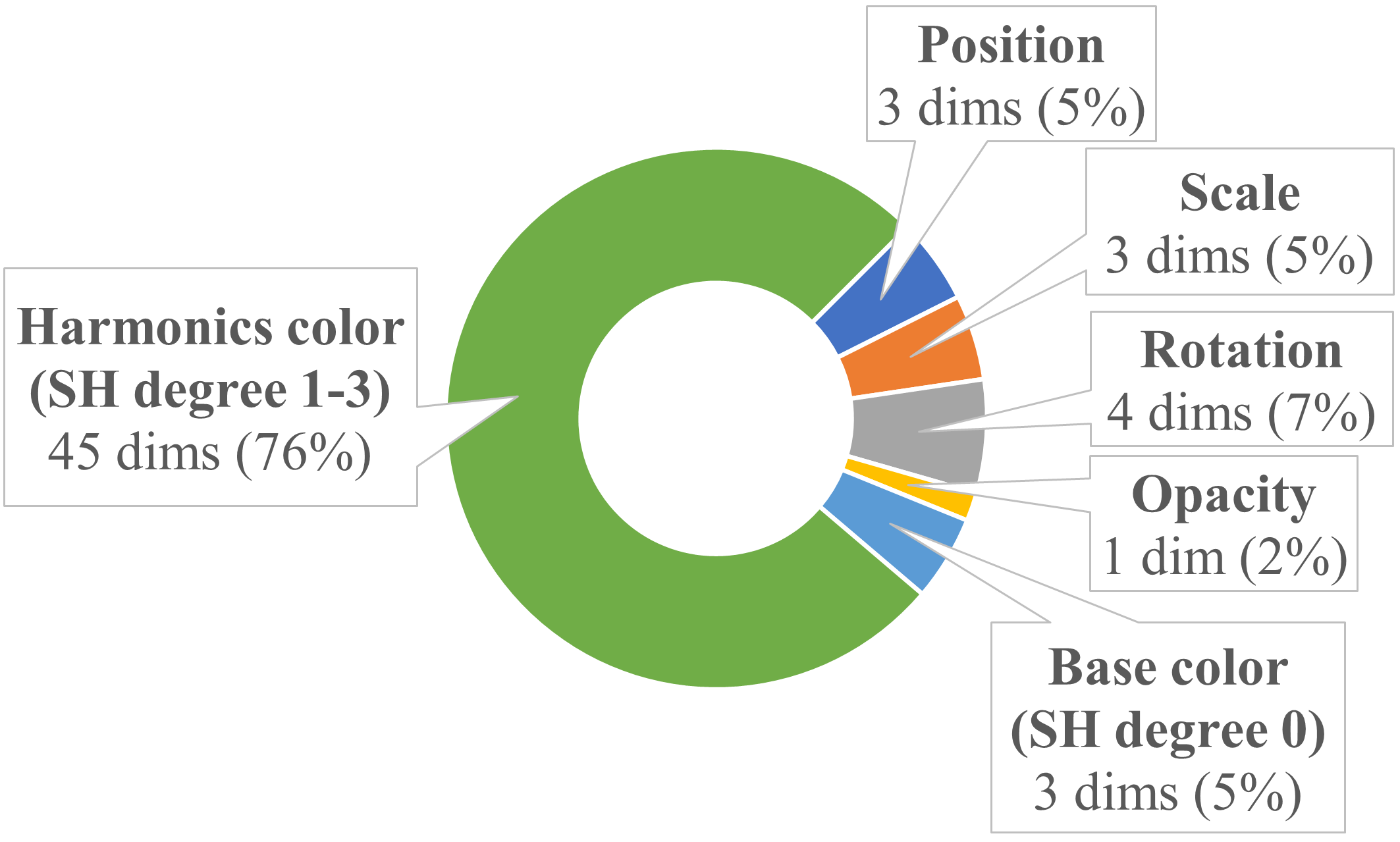}
  \caption{Composition of each Gaussian. Dimensions of each attribute and its percentage in size are marked in the pie chart.}
  \label{fig:composition}
\end{wrapfigure}

As a promising scene representation method, 3D Gaussian Splatting (3DGS) has demonstrated its powerful ability on novel view synthesis~\cite{kerbl20233d}. Different from NeRF which implicitly stores scene properties based on neural networks~\cite{mildenhall2021nerf}, 3DGS represents the underlying geometry and appearance explicitly using the anisotropic ellipsoid, which is parameterized by position $\mathbf{\mu} \in \mathbb{R}^3$, covariance $\mathbf{\Sigma} \in \mathbb{R}^{3\times3}$, color $\textbf{c} \in \mathbb{R}^3$, and opacity $\alpha \in \mathbb{R}$. Covariance $\mathbf{\Sigma}$ can be further factorized into scale vector $s\in\mathbb{R}^3$ and rotation vector $r\in\mathbb{R}^4$. To render a novel view, 3DGS adopts a tile-based differentiable rasterization, where 3D Gaussians are projected onto the camera plane and thus form 2D Gaussians, followed by a point-based rendering operation which is formulated as follows:
\begin{equation}
\centering
\begin{split}
    M(\mathbf{p},\mathbf{\mu}_u, \mathbf{\Sigma}_u) =& e^{-\frac{1}{2}((\mathbf{p}-\mathbf{\mu}_u)^T\mathbf{\Sigma}_u^{-1}(\mathbf{p}-{\mu}_u))},\\
    \mathbf{c(p)} =& \sum_{u \in \mathcal{Q}} \mathbf{c}_u\sigma_u\prod_{v=1}^{u-1}(1-\sigma_v), \sigma_u = \alpha_uM(\mathbf{p},\mathbf{\mu}_u, \mathbf{\Sigma}_u),
\end{split}
\end{equation}
where $\mathbf{p}$ is the target pixel, $\mathbf{c}(\mathbf{p})$ is the rendered color of $\mathbf{p}$, $\mathbf{\mu}_u, \mathbf{\Sigma}_u, \alpha_u$ represent 2D Gaussian position, covariance and opacity projected from the $u$-th 3D Gaussian, $\mathcal{Q}$ means the number of 3D Gaussian located inside this tile. By means of GPU-efficient optimization, impressive real-time rendering can be obtained.

However, 3DGS's high rendering quality and speed are at the expense of substantial storage capacity. As illustrated in Fig.~\ref{fig:composition}, each Gaussian is characterized by $59$ \texttt{float32} numbers. Among these parameters, $45$ (approximately $76\%$) pertain to Spherical Harmonics (SHs) of degrees $1$-$3$. Despite comprising the majority of Gaussian parameters, these SHs contribute slightly to rendering quality as demonstrated in Fig.~\ref{fig:shprune}. Therefore, removing the redundant parameters while retaining the view-dependent appearance is necessary.

\subsection{Rate-Distortion Optimized 3D Gaussian}
\label{sec:rdo_gaussian}

\subsubsection{Overview.}

As depicted in Fig.~\ref{fig:pipeline}, our optimization process comprises four main components: Gaussian pruning, adaptive spherical harmonics (SHs) pruning, entropy-constrained vector quantization (ECVQ), and rendering. To begin with, the input Gaussians are pruned by learned Gaussian masks. Subsequently, adaptive SH masks are applied to the SHs of each Gaussian, allowing that different Gaussians can have different degrees of SHs. Furthermore, covariance and color parameters are quantized by ECVQ to obtain a more compact representation, followed by a rendering process that is executed on the pruned and quantized 3D Gaussians.

\subsubsection{Gaussian Pruning.}

To accurately represent complex scenes, 3DGS introduces a mechanism called adaptive density control which adjusts the number of Gaussians during the training process~\cite{kerbl20233d}. Specifically, adaptive density control first identifies regions with missing geometric features (under-reconstruction) as well as regions with large Gaussians (over-reconstruction), where Gaussians are either cloned or split to rectify these regions. In addition, Gaussians with low opacity are periodically removed to prevent floaters close to cameras. While adaptive density control can significantly enhance rendering quality, millions of Gaussians will be generated during this process and thus lead to substantial memory costs.

Inspired by Compact-3DGS~\cite{lee2023compact}, we prune the Gaussians based on scales and opacities with the help of learnable masks, since Gaussians with small scales and low opacities contribute little to the overall rendering quality in most circumstances. Specifically, we assign a mask parameterized by $\phi_i\in\mathbb{R}$ to each Gaussian, indicating whether it is pruned or not, where $i$ represents the Gaussian index. The mask value is mapped to $(0,1)$ using the \texttt{sigmoid} function, which is denoted as the soft mask:
\begin{equation}
    \phi_i^{\rm soft} = \texttt{sigmoid}\left(\phi_i\right), \ \phi_i^{\rm soft} \in \left(0,1\right).
\end{equation}
The soft mask is binarized by a predefined threshold $\phi_{\rm thres}$. Since the binarization process is not differentiable, we employ the straight-through estimator (STE)~\cite{bengio2013estimating} to bypass its gradient, resulting in the hard mask:
\begin{equation}
    \phi_i^{\rm hard} = \texttt{sg}\left(\mathbbm{1}\left(\phi_i^{\rm soft} > \phi_{\rm thres}\right) - \phi_i^{\rm soft}\right) + \phi_i^{\rm soft}, \ \phi_i^{\rm hard} \in \left\{0,1\right\},
\end{equation}
where $\mathbbm{1}$ is the indicator function and \texttt{sg} denotes the stop gradient operation.

The hard mask is then applied to the scale and opacity attributes before rendering. Specifically, if one Gaussian is marked as pruned, its scale and opacity become zero; otherwise, they remain intact. This process is formulated as follows:
\begin{equation}
    \mathbf{\hat{s}}_i = \phi_i^{\rm hard}\mathbf{s}_i, \ \hat{\alpha}_i = \phi_i^{\rm hard}\alpha_i,
\end{equation}
where $\mathbf{s}_i\in\mathbb{R}^3$, $\alpha_i\in\mathbb{R}$ represent the scale and opacity values of the $i$-th Gaussian respectively, and $\mathbf{\hat{s}}_i$, $\hat{\alpha}_i$ denote their masked versions.

Since both soft masks and hard masks are designed to reduce the number of Gaussians, only when their values are optimized below the threshold can they take effect. Hence, we define the training objective of Gaussian masks as follows:
\begin{equation}
    \mathcal{L}_{\rm GSprune} = \frac{1}{N}\sum_i\phi_i^{\rm soft},
\end{equation}
where $N$ is the total number of Gaussians.

It is important to note that after one training iteration, though some Gaussians are labeled as pruned by the masks, they are not actually removed. The masks are dynamic, meaning that their values are continuously optimized until training ends. Once they surpass the threshold, the corresponding Gaussians will respawn. Upon completion of training, these Gaussians are finally eliminated and not stored anymore.

\subsubsection{Adaptive Spherical Harmonics Pruning.}

Following prior work\cite{fridovich2022plenoxels}, 3DGS employs Spherical Harmonics (SHs) to represent the color of the radiance field. By default, SH parameters extend up to $4$ degrees: the degree-0 or DC component determines the base diffuse color, while degree-1 to degree-3 components manage view-dependent effects such as reflection. Though SHs of higher degrees are able to capture a more intricate view-dependent appearance, with increasing SH degrees, the number of SH parameters also grows exponentially.
Considering that scene complexity varies across different views, the SH parameters of each Gaussian need to be adaptive based on scene complexity, instead of treating the degrees of each Gaussian equally.

To this end, we design a degree-adaptive mask pruning strategy to eliminate redundant SH parameters. The mask is end-to-end optimized to automatically learn the complex view-dependent effects and diffused regions. Similar to Gaussian pruning, each Gaussian is assigned $3$ learnable masks corresponding to the $3$ harmonic degrees:
\begin{equation}
\begin{split}
    \theta_i^{(l){\rm soft}} &= \texttt{sigmoid}\left(\theta_i^{(l)}\right), \ \theta_i^{(l){\rm soft}} \in \left(0,1\right), \\
    \theta_i^{(l){\rm hard}} &= \texttt{sg}\left(\mathbbm{1}\left(\theta_i^{(l){\rm soft}} > \theta_{\rm thres}\right) - \theta_i^{(l){\rm soft}}\right) + \theta_i^{(l){\rm soft}}, \ \theta_i^{(l){\rm hard}} \in \left\{0,1\right\},
\end{split}
\end{equation}
where $\theta_i^{(l)}$ represents the raw mask value for degree-$l$ of the $i$-th Gaussian, $\theta_i^{(l){\rm soft}}$ and $\theta_i^{(l){\rm hard}}$ denote its soft and hard versions respectively. Before rendering, the hard SH mask is applied to the corresponding degree of SH parameters, setting them to zero if the mask value is zero and unchanged otherwise:
\begin{equation}
    \mathbf{\hat{c}}_i^{(l)} = \theta^{{(l)\rm hard}}\mathbf{c}_i^{(l)},
\end{equation}
where $\mathbf{c}_i^{(l)}\in\mathbb{R}^{(2l+1)\times3}$ is the degree-$l$ SH coefficients of the $i$-th Gaussian and $\mathbf{\hat{c}}_i^{(l)}$ denotes its masked version.

Given the varying number of SH coefficients between degrees, the SHs pruning loss of each degree is weighted by its number of coefficients:
\begin{equation}
    \mathcal{L}_{\rm SHprune} = \frac{1}{N}\sum_i\sum_{l=1}^k\frac{2l+1}{(k+1)^2-1}\theta_i^{(l){\rm soft}},
\end{equation}
where $k$ denotes the maximum degree of SHs ($k=3$ by default).

\subsubsection{Entropy-Constrained Vector Quantization.}
By means of Gaussian pruning and adaptive SHs pruning, the number of Gaussians and SH parameters can be significantly reduced. Despite these efforts, there are still some redundancies that persist within the attributes of Gaussians. As illustrated in Fig.~\ref{fig:composition}, these attributes are usually high-dimensional vectors with numerous bits to represent, which motivates us to convert them into a compact discrete representation.

Vector quantization (VQ)~\cite{gersho2012vector} is a prevalent technique used in common compression tasks. For continuous vector inputs, VQ identifies the most suitable representative codeword vector from the codebook and substitutes the vector with the corresponding codeword index. Consequently, VQ transforms a continuous vector into a discrete index, resulting in significant bit savings. Considering the superiority of VQ, most 3DGS compression works~\cite{navaneet2023compact3d,lee2023compact,fan2023lightgaussian} adopt VQ as the main part of their compression procedures. However, during the step of codeword selection, existing works only choose the codeword with minimal distortion (\ie, L2 distance between input vector and codeword vector) regardless of the rate, resulting in sub-optimal rate-distortion (RD) performance. In pursuit of RD optimization, we adopt ECVQ~\cite{chou1989entropy} as our quantization approach, which considers both rate and distortion in codeword selection and can ensure better RD performance.

In order to estimate the rate of each codeword, we devise a discrete unconditional entropy model $P$ to estimate the probability distribution of each index. The probability of index $j$ in the codebook is denoted as $p_j$, satisfying $\sum_{j=1}^M p_j=1$ and $p_j>0$. However, directly optimizing $p_j$ forms a constrained optimization problem, which is hard to solve. To optimize $p_j$ unconstrainedly, we parameterize $p_j$ with the \texttt{softmax} function and unnormalized logits $w=(w_1,w_2, ...,w_M)$:
\begin{equation}
    p_j = \frac{e^{-w_j}}{\sum_{m=1}^M e^{-w_m}},
\end{equation}
where $M$ is the codebook size.

Among all Gaussian attributes, we quantize three of them: scale, rotation and color. For the color attribute, since adaptive SHs pruning precedes ECVQ, the degrees of each Gaussian may vary. To avoid zero-value appearing in codebooks, four codebooks are allocated for all $4$ degrees of SHs, resulting in $6$ codebooks in total. Taking the scale attribute as an example, given a scale tensor $\mathbf{s}_i$, we select the index $j$ with minimal RD loss:
\begin{equation}
    j = \mathop{\arg\min}\limits_{m} \frac{r_{i,m}^{(s)}}{\lambda^{(s)}}+d_{i,m}^{(s)} = \mathop{\arg\min}\limits_{m} -\frac{\log p_m}{\lambda^{(s)}} + \left(\mathbf{s}_i - {\rm CB}^{(s)}[m]\right)^2,
\end{equation}
where ${\rm CB}^{(s)}$ is the codebook for scale, $[\cdot]$ is the indexing operation, $r_{i,m}^{(s)}$ is the rate loss if ${\rm CB}^{(s)}[m]$ is selected as the codeword for $\mathbf{s}_i$, $d_{i,m}^{(s)}$ is the VQ codeword loss respectively and $\lambda^{(s)}$ is the hyperparameter controlling the balance between rate and distortion.

We sum up the rate loss and VQ loss of every Gaussian attribute and average on all Gaussians to obtain the total rate loss and VQ loss:
\begin{equation}
\begin{split}
    \mathcal{L}_{\rm rate} &= \frac{1}{N}\sum_{i=1}^N \frac{r_{i,j}^{(s)}}{\lambda^{(s)}}
    + \frac{r_{i,j}^{(r)}}{\lambda^{(r)}}
    + \frac{r_{i,j}^{(DC)}}{\lambda^{(DC)}}
    + \frac{r_{i,j}^{(SH1)}}{\lambda^{(SH1)}}
    + \frac{r_{i,j}^{(SH2)}}{\lambda^{(SH2)}}
    + \frac{r_{i,j}^{(SH3)}}{\lambda^{(SH3)}}, \\
    \mathcal{L}_{\rm VQ} &= \frac{1}{N}\sum_{i=1}^N d_{i,j}^{(s)} + d_{i,j}^{(r)} + d_{i,j}^{(DC)} + d_{i,j}^{(SH1)} + d_{i,j}^{(SH2)} + d_{i,j}^{(SH3)},
\end{split}
\end{equation}
where superscripts $(s)$, $(r)$, $(DC)$, $(SH1)$, $(SH2)$, $(SH3)$ denote scale, rotation, DC component of color, and degree $1$-$3$ harmonics of color, respectively.

\subsubsection{Training Objective.}

Following ECVQ, we obtain the quantized Gaussians for rendering. The rendering loss combines both L1 loss and SSIM loss with $\lambda_{\rm SSIM}$ balancing their weights, which is the same as 3DGS~\cite{kerbl20233d}:
\begin{equation}
    \mathcal{L}_{\rm render} = \left(1-\lambda_{\rm SSIM}\right)\mathcal{L}_1 + \lambda_{\rm SSIM}\mathcal{L}_{\rm D-SSIM}.
\end{equation}
Finally, the losses generated during the above process are summed up with the rendering loss with respective weights:
\begin{equation}
    \label{eq: total_loss}
    \mathcal{L}_{\rm total}
    = \lambda_{\rm GSprune}\mathcal{L}_{\rm GSprune}
    + \lambda_{\rm SHprune}\mathcal{L}_{\rm SHprune}
    + \mathcal{L}_{\rm rate}
    + \mathcal{L}_{\rm VQ}
    + \mathcal{L}_{\rm render}
\end{equation}

\subsection{Parameters Encoding}
\label{sec:encoding}

\label{subsec:parameter encoding}
After optimization, we acquire a pruned and quantized Gaussian representation, which can be further encoded for storage and transmission purposes.

\subsubsection{Gaussians Removement.}
While many Gaussians are marked as pruned by the Gaussian pruning mask, they are not actually removed and continue being optimized during compression-aware training. After training, they can be completely removed and not stored anymore.

\subsubsection{Gaussians Rearrangement.}
\label{subsubsec: rearranging gaussians}
During the adaptive SHs pruning step, SH parameters of Gaussians are pruned to various degrees. If they are directly removed as the previous step does, it's not possible to locate where the remaining parameters are when rendering. As a result, we need to pack SH masks into the representation to mark the pruned locations, which needs additional storage costs.

Fortunately, the Gaussian representation is order-invariant where rearranging Gaussians does not impact the rendering results. We can cluster Gaussians with the same SH masks and reorder Gaussians by them. Since the position of SH parameters are identical in the same cluster, SH masks are not needed anymore and the only storage overhead is the indexes where a kind of SH masks first appears. Specifically, we treat the $3$ bits SH hard masks as the integer (e.g., SH mask $011$ corresponds to integer $3$) and sort the Gaussians based on this integer. Following sorting, the index where each integer first appears is stored as the starting position of a specific kind of mask. Therefore, only $8$ additional indexes need to be stored.

\subsubsection{Arithmetic Coding.}
Attributes undergoing ECVQ can be represented by integer indexes. Leveraging the unconditional entropy model learned during the ECVQ process, we obtain the probability distribution of each index. These distributions are utilized in arithmetic coding to eliminate the statistical redundancies between attribute values. In addition to the encoded indexes, codebooks and logits also need to be stored for decoding. Note that codewords not used by any Gaussian as well as the corresponding logits are removed to reduce the sizes of the codebook and logits.

\subsubsection{Non-VQ Attributes.}
Two attributes, position and opacity, are not quantized. Previous research indicates that quantizing positions may lead to overlapping Gaussians~\cite{navaneet2023compact3d}, significantly impacting performance. Therefore, we simply reduce its precision from \texttt{float32} to \texttt{float16} to decrease its size without affecting performance. Regarding opacity, being a scalar rather than a vector, we opt for scalar quantization and map its value to an $8$-bit integer.

\section{Experiments}
\label{sec:experiments}

\subsection{Experimental Settings}

\subsubsection{Datasets.}
 We evaluate our proposed method on four commonly-adopted datasets: the Mip-NeRF360 dataset~\cite{barron2022mip}, the Tanks and Temples dataset~\cite{knapitsch2017tanks}, the Deep Blending dataset~\cite{hedman2018deep} and the NeRF-Synthetic dataset~\cite{mildenhall2021nerf}. For the former three datasets, various real-world scenes that contain both indoor and outdoor scenes are selected. For the NeRF-Synthetic dataset, eight 360$^{\circ}$ object-centric scenes with black background are utilized. We follow the experimental protocols provided by 3DGS~\cite{kerbl20233d}.

\subsubsection{Implementation Details.}
We first train the Gaussians based solely on 3DGS~\cite{kerbl20233d} up to 15k iterations. Subsequently, we introduce our rate-distortion optimization, incorporating Gaussian pruning, adaptive SHs pruning and ECVQ into the optimization pipeline. During this process, parameters generated from the aforementioned operations are jointly optimized with classical Gaussian parameters (\ie, parameters illustrated in Sec.~\ref{sec:3dgs}) until the end of training. After this optimization process, we pack the trained Gaussian radiance field to the final representation using the method mentioned in Sec.~\ref{subsec:parameter encoding}. Kindly refer to the supplementary materials for more details on the hyperparameters used.

\subsubsection{Baselines}
\label{subsubsec: baselines}
We compare the performace of our proposed RDO-Gaussian with 3DGS~\cite{kerbl20233d}, CompGS~\cite{navaneet2023compact3d}, Compact-3DGS~\cite{lee2023compact} and LightGaussian~\cite{fan2023lightgaussian}.For a fair comparisom, We reimplemented 3DGS by ourselves. The performance of CompGS is computed from their pre-trained models, and the performance of LightGaussian and Compact-3DGS are quoted from their papers. 

\subsection{Results}

\begin{figure}[tb]
  \centering
  \includegraphics[width=1\textwidth]{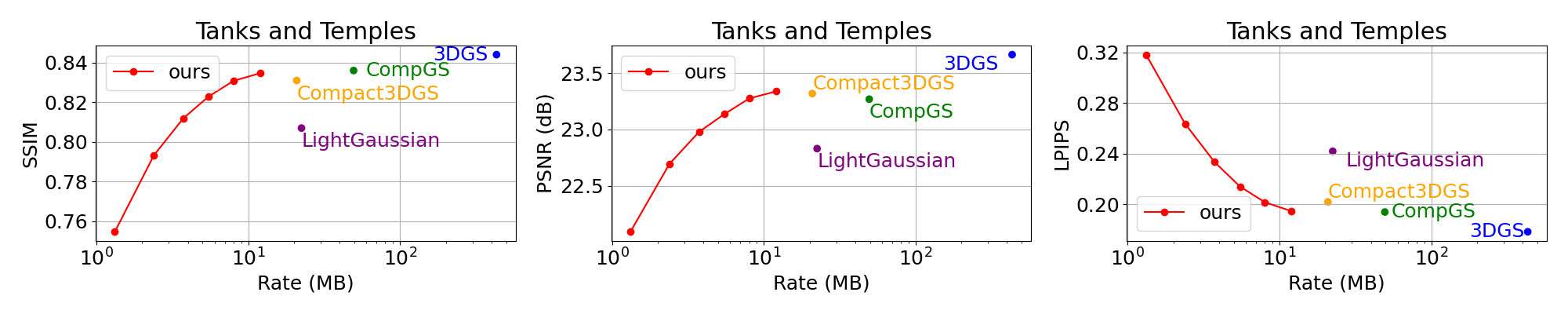}
  \includegraphics[width=1\textwidth]{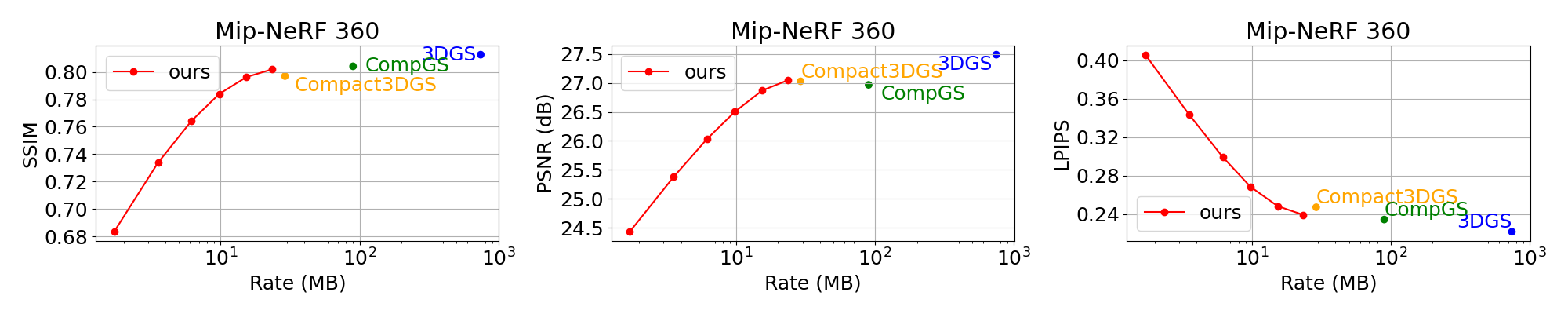}
  \includegraphics[width=1\textwidth]{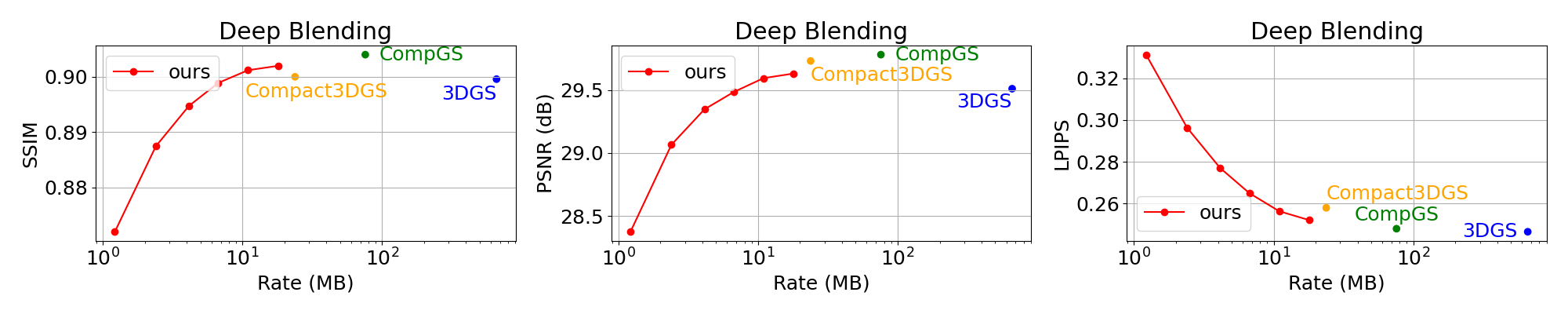}
  \includegraphics[width=1\textwidth]{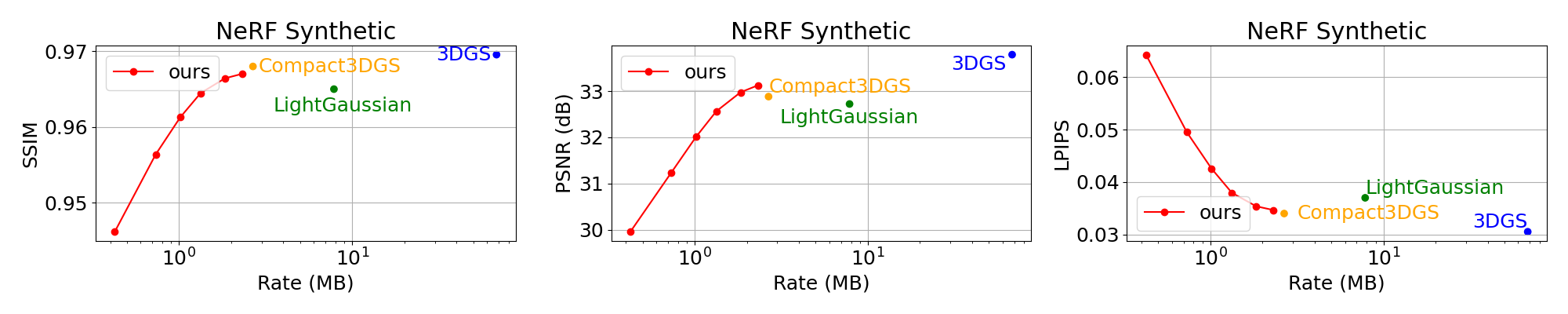}
  
  \caption{Rate-Distortion curves of our method on all four datasets. On the x-axis is the rate of the compressed Gaussian representation (in logarithmic scale) and the y-axis represents the corresponding quality metric (SSIM, PSNR, LPIPS). Both are averaged over scenes in a dataset. We also mark the performance of 3DGS (implemented by ourselves) and other 3DGS compression works (from pre-trained models or papers).}
  \label{fig:rdcurve}
\end{figure}

\subsubsection{Rate-Distortion Performance.}

By modifying the $\lambda_{\rm SHprune}$ and $\lambda_{\rm GSprune}$ in Eq.~\ref{eq: total_loss}, we obtain a series of compressed Gaussians with different sizes. Novel views rendered from these Gaussians are measured and averaged using metrics such as PSNR, SSIM and LPIPS, where a rate-distortion (RD) curve is drawn in Fig.~\ref{fig:rdcurve}. Moreover, we also present the performance of 3DGS and other 3DGS compression works mentioned in Sec.~\ref{subsubsec: baselines} on the RD curve, where Gaussians of only one single size exist.

According to Fig.~\ref{fig:rdcurve}, it is observed that compared to 3DGS, our proposed method is able to achieve over $40$$\times$ compression ratio without severe degradation of quality. If we further decrease the rates, we can compress Gaussians at over $300$$\times$ compression ratio with approximately $1$ MB size, while a typical JPEG image in Mip-NeRF360 dataset~\cite{barron2022mip} with resolution $3000\times2000$ still needs a storage space of $3$ MB. Such an observation indicates that our method can compress Gaussian to a size less than an image with adequate scene information, which is important in extremely band-limited circumstances. Compared with existing Gaussian compression methods, RDO-Gaussian can also achieve over $30\%$ bitrate savings on average.

\subsubsection{Quantitative Comparisons.}

\begin{figure}[tb]
  \centering
  \includegraphics[width=1\textwidth]{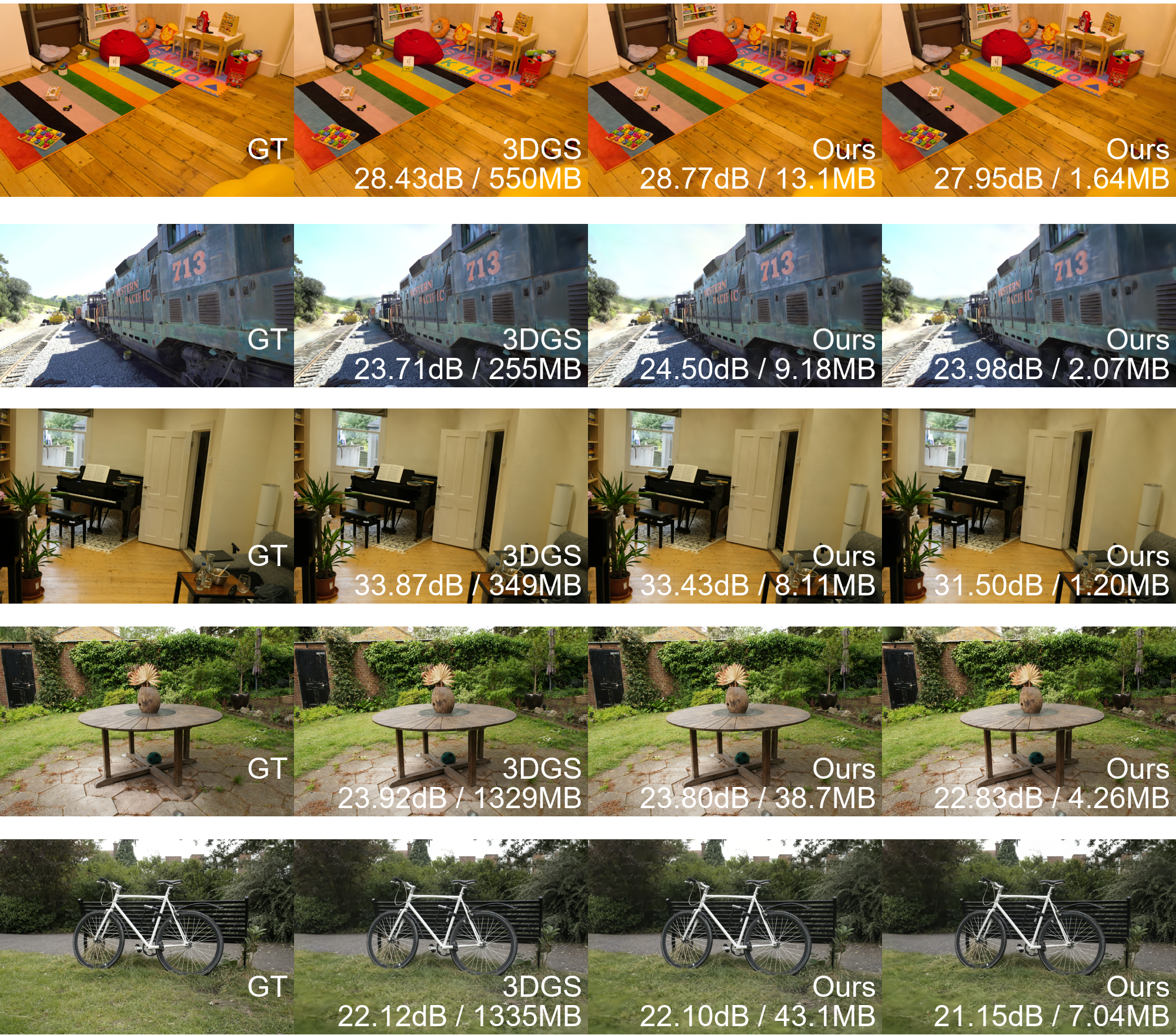}
  \caption{Perceptual comparison of our method and 3DGS. We extract the rendered images from our Gaussian representations of different rates. The PSNR values and rates are marked in each rendered image. The average compression ratio can be up to $40$$\times$ for the high-rate representation and $300$$\times$ for the low-rate representation.}
  \label{fig:perceptual}
\end{figure}

To validate the performance of our representation at different RD trade-offs, we showcase the images rendered from our high-rate and low-rate Gaussian representations respectively. As shown in Fig.~\ref{fig:perceptual}, our high-rate representation achieves over $40$$\times$ compression ratio on average while maintaining the same quality as 3DGS. For our low-rate representation, Gaussians can be compressed to the size of an image ($1\sim3$ MB) without severe distortions.

\subsection{Ablation Studies}

\subsubsection{Bitrate Savings of Each Module.}

To evaluate the bitrate savings achieved by each module, we progressively incorporate these modules into 3DGS. Subsequently, we measure changes in the rate and PSNR across different scenes, as depicted in Tab.~\ref{tab:bitrate savings}. In scene Truck, Gaussian pruning reduces around $55\%$ Gaussians with a minor PSNR drop. Subsequent adaptive SHs pruning eliminates redundant SH parameters for each Gaussian, saving approximately $65\%$ bitrate compared to the previous one. Following pruning, ECVQ discretizes continuous Gaussian attributes (excluding position and opacity), resulting in substantial bitrate savings albeit with most performance degradation. Lastly, parameter compression reduces around half the bitrate. These four steps combined achieve over $43\times$ compression ratio. For scene Playroom, the PSNR increases as the rate goes down, contrary to the expected decrease. This anomaly arises from 3DGS overfitting to training images, while pruning and quantization techniques inherently mitigate overfitting issues.

\begin{table}[tb]
  \caption{Bitrate savings achieved through various techniques. Beginning with 3DGS\cite{kerbl20233d}, we incrementally introduce our methodologies to quantify the reduction in bitrate compared to the preceding step.}
  \label{tab:bitrate savings}
  \centering
  \begin{tabular}{@{}l@{\hspace{1mm}}|@{\hspace{1mm}}c@{\hspace{1mm}}|@{\hspace{1mm}}c@{\hspace{1mm}}|@{\hspace{1mm}}c@{\hspace{1mm}}|@{\hspace{1mm}}c@{\hspace{1mm}}|@{\hspace{1mm}}c@{\hspace{1mm}}|@{\hspace{1mm}}c@{}}
    \toprule
    & \multicolumn{3}{c@{\hspace{1mm}}|@{\hspace{1mm}}}{Truck} & \multicolumn{3}{c}{Playroom} \\
    \cmidrule{2-7}          Module       & Rate (MB) & Savings (\%) & PSNR  & Rate (MB) & Savings (\%) & PSNR  \\
    \midrule
    3DGS            & $609.3$   &      -      & $25.36$ & $550.5$ &     -       & $29.95$ \\
    +GS prune       & $275.2$   &  $-54.83\%$ & $25.34$ & $252.5$ & $-54.14\%$  & $29.97$ \\
    +Ada. SHs prune & $ 94.1$   &  $-65.81\%$ & $25.32$ & $ 81.9$ & $-67.58\%$  & $30.04$ \\
    +ECVQ           & $ 33.9$   &  $-63.98\%$ & $25.06$ & $ 27.3$ & $-66.62\%$  & $30.18$ \\
    +Param comp     & $ 14.9$   &  $-56.15\%$ & $25.04$ & $ 13.1$ & $-52.02\%$  & $30.16$ \\
  \bottomrule
  \end{tabular}
\end{table}

\subsubsection{Effect of Adaptive SHs Pruning.}

\begin{figure}[tb]
  \centering
  % Materials
  \begin{subfigure}{0.193\linewidth}
    \includegraphics[width=1\textwidth]{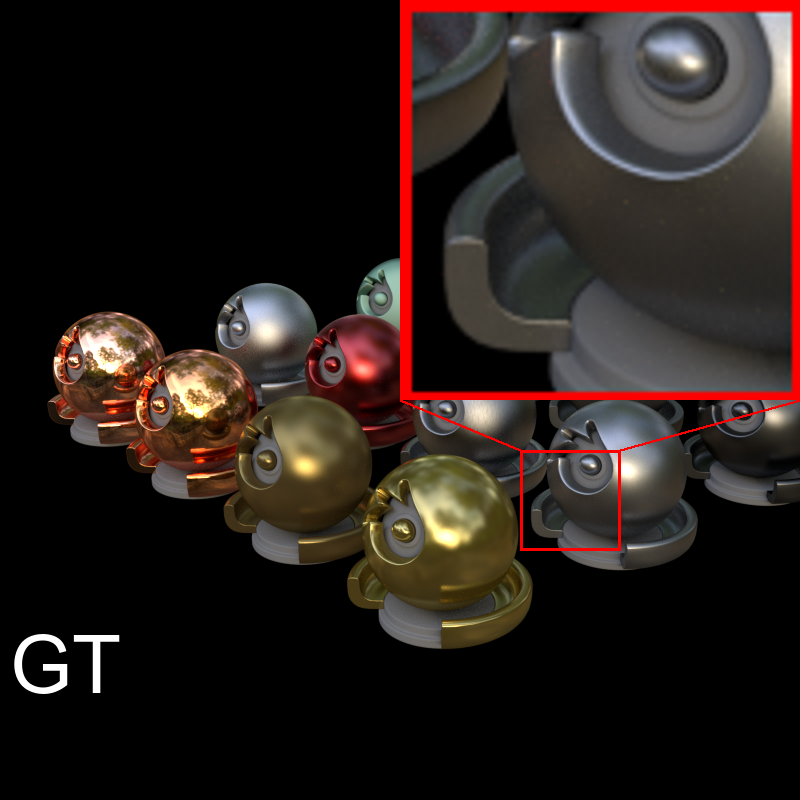}
    \label{fig:materials_gt}
  \end{subfigure}
  \hfill
  \begin{subfigure}{0.193\linewidth}
    \includegraphics[width=1\textwidth]{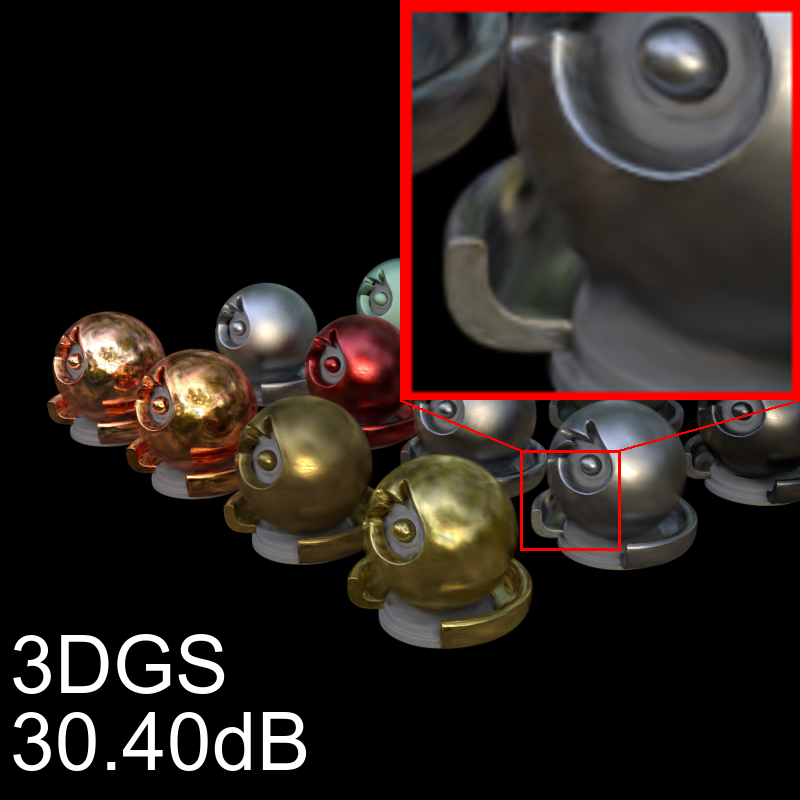}
    \label{fig:materials_gs}
  \end{subfigure}
  \begin{subfigure}{0.193\linewidth}
    \includegraphics[width=1\textwidth]{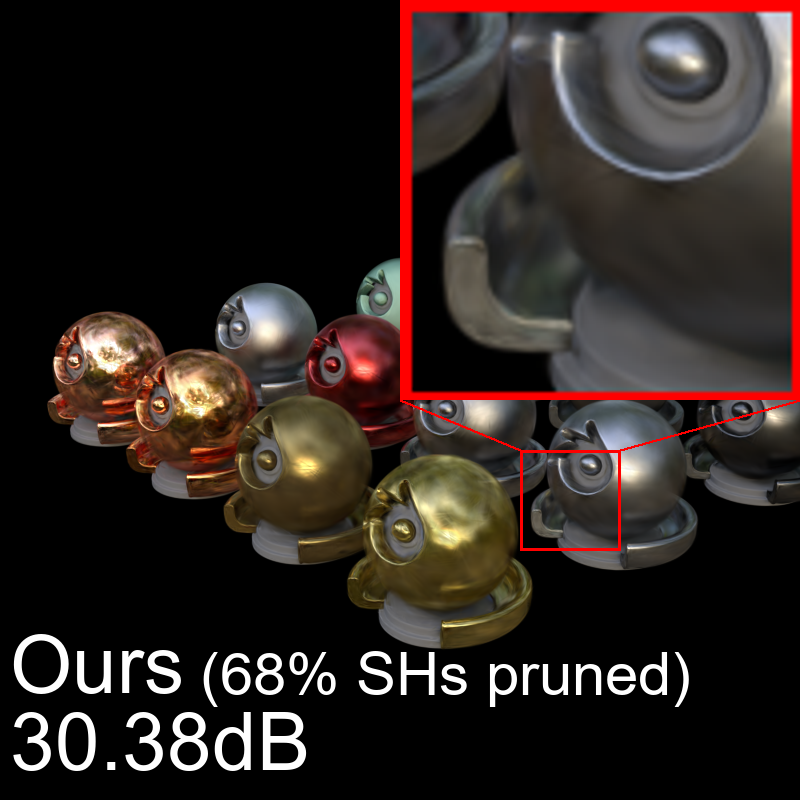}
    \label{fig:materials_ours0.68}
  \end{subfigure}
  \begin{subfigure}{0.193\linewidth}
    \includegraphics[width=1\textwidth]{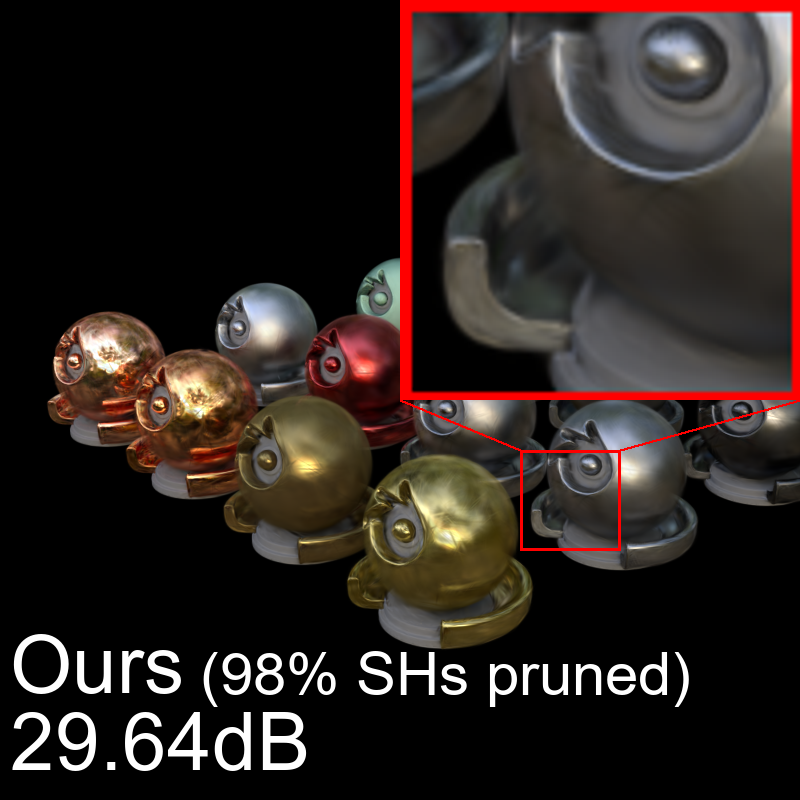}
    \label{fig:materials_ours0.98}
  \end{subfigure}
  \begin{subfigure}{0.193\linewidth}
    \includegraphics[width=1\textwidth]{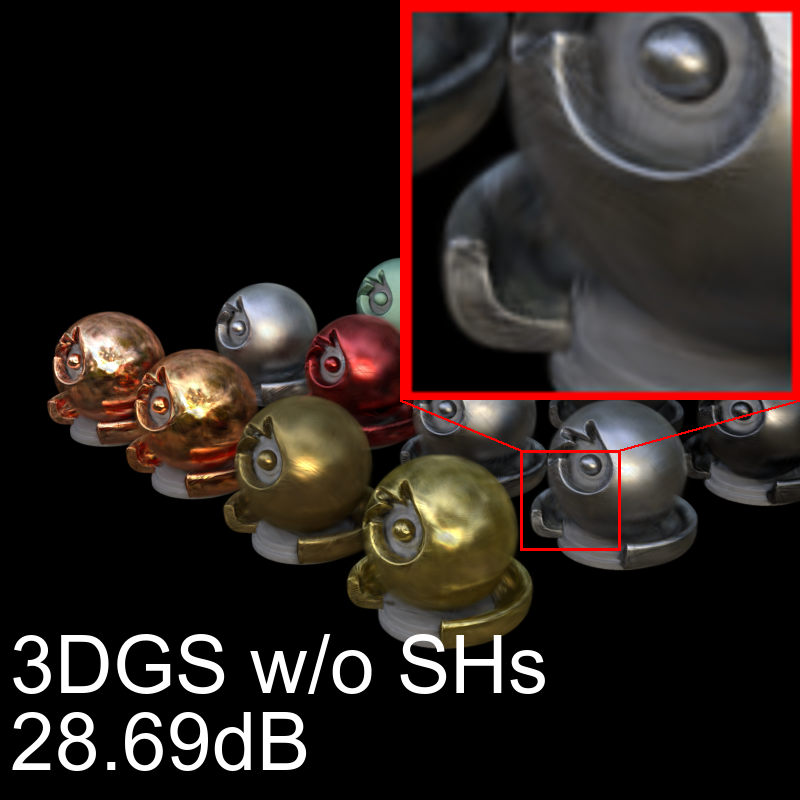}
    \label{fig:materials_nosh}
  \end{subfigure}
  
  % Drums
  \begin{subfigure}{0.193\linewidth}
    \includegraphics[width=1\textwidth]{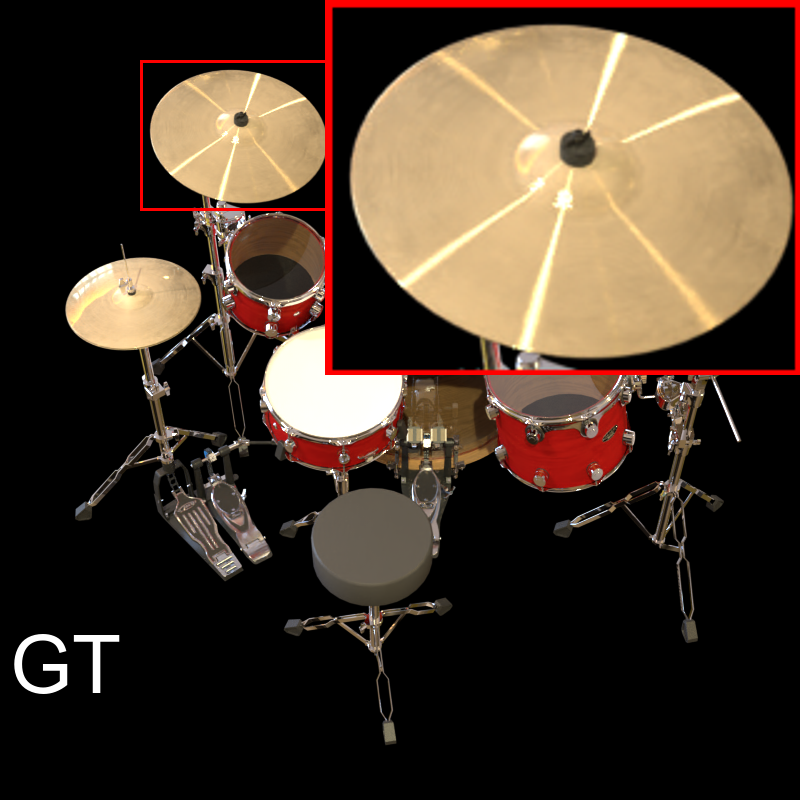}
    \label{fig:drum_gt}
  \end{subfigure}
  \hfill
  \begin{subfigure}{0.193\linewidth}
    \includegraphics[width=1\textwidth]{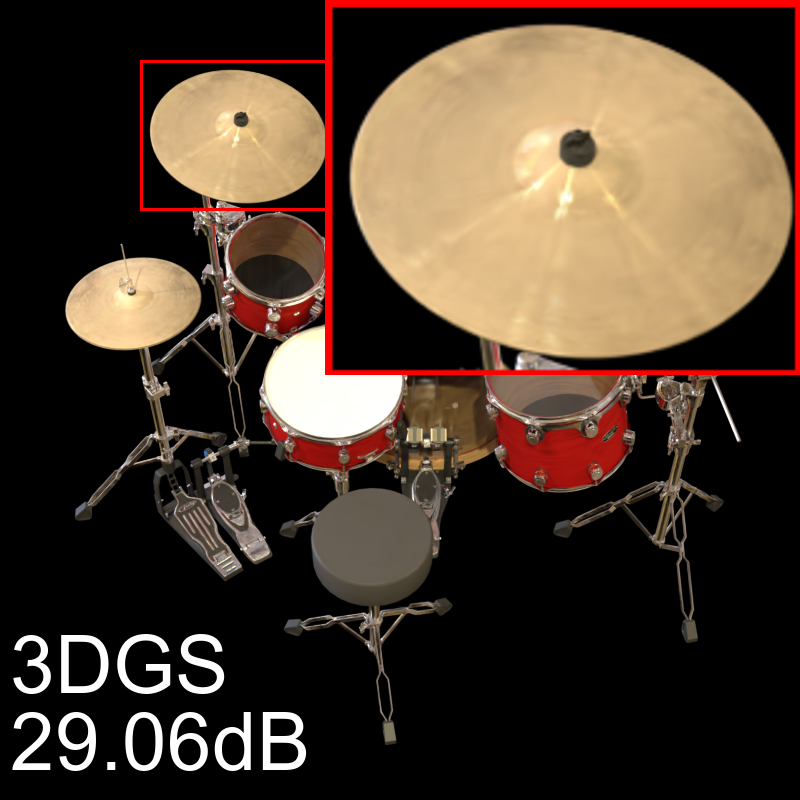}
    \label{fig:drum_gs}
  \end{subfigure}
  \begin{subfigure}{0.193\linewidth}
    \includegraphics[width=1\textwidth]{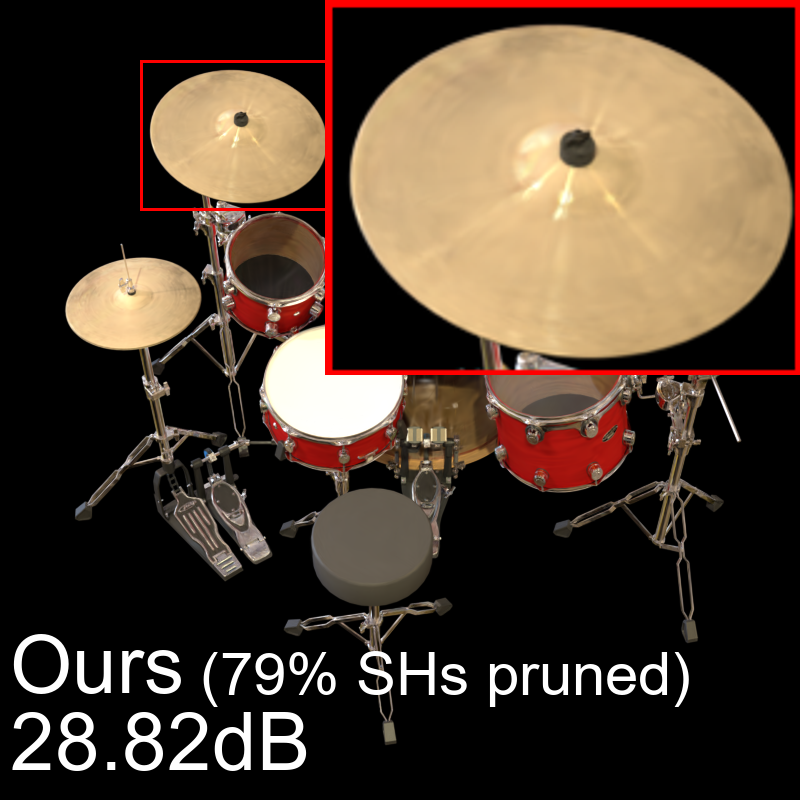}
    \label{fig:drum_ours0.79}
  \end{subfigure}
  \begin{subfigure}{0.193\linewidth}
    \includegraphics[width=1\textwidth]{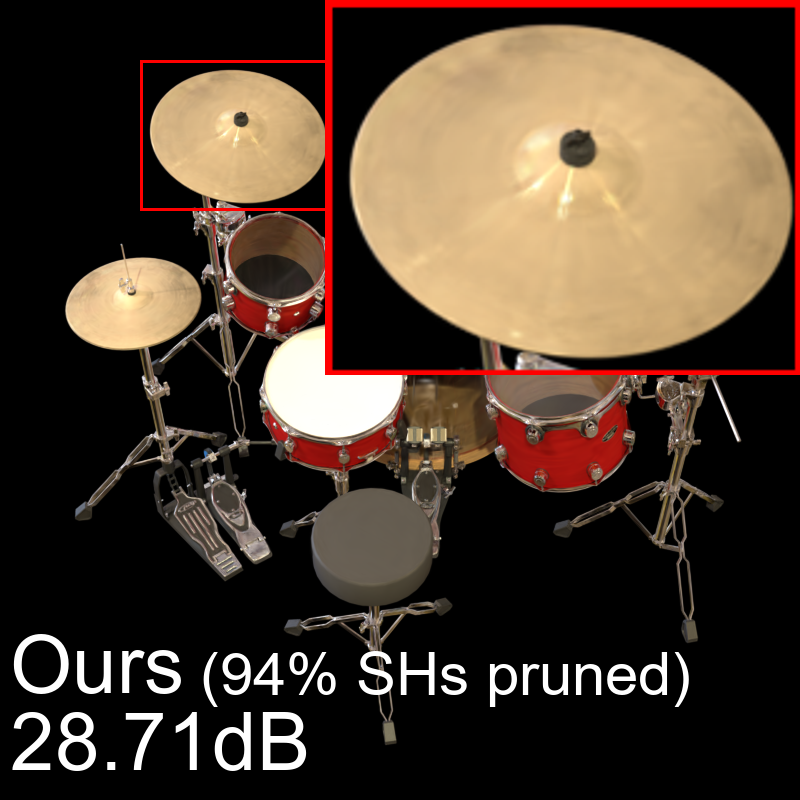}
    \label{fig:drum_ours0.94}
  \end{subfigure}
  \begin{subfigure}{0.193\linewidth}
    \includegraphics[width=1\textwidth]{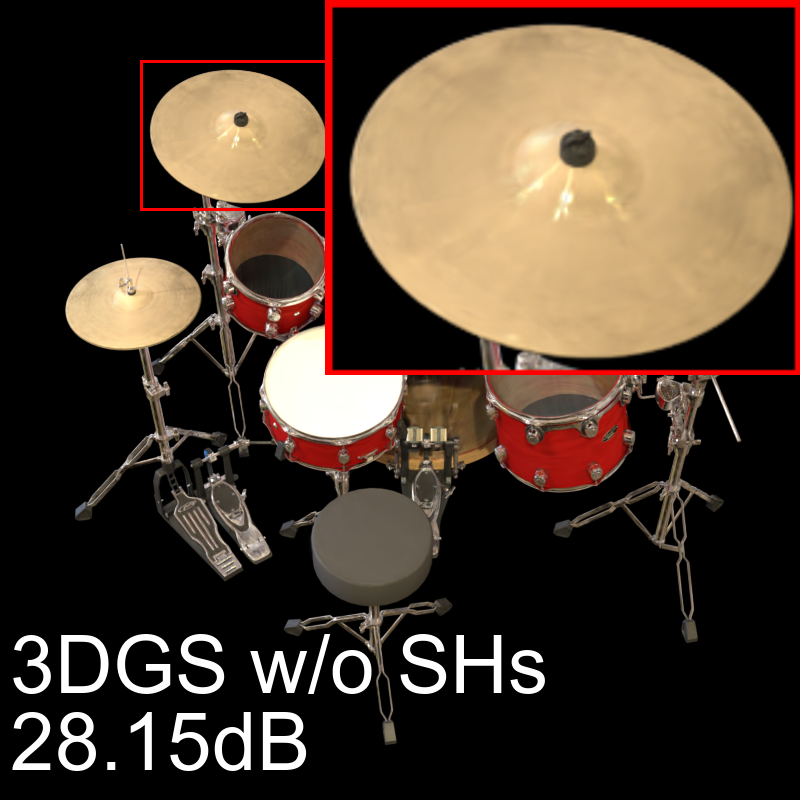}
    \label{fig:drum_nosh}
  \end{subfigure}
  
  \caption{Effect of our adaptive SHs pruning. Our method can preserve most view-dependent effects modeled by 3DGS with minor PSNR drops while pruning abundant SH parameters.}
  \label{fig:shprune}
\end{figure}

To further validate the effect of adaptive SHs pruning and its volumetric bit allocation function, we exclusively maintain adaptive SH pruning and deactivate other modules during training. We evaluate our method across various SH pruning ratios on scenes Drums and Materials from the NeRF-Synthetic dataset\cite{mildenhall2021nerf}. For comparison, we present images from 3DGS and 3DGS without SHs, where the maximum SH parameter degree is restricted to 0.

As shown in Fig.~\ref{fig:shprune}, we observe that view-dependent effects such as specular reflection are evident in both scenes. Our method can prune redundant SH parameters while retaining view-dependent effects modeled by 3DGS, such as the reflection on drums and edges of the ball. When pruning ratios are around 70\%, the PSNR drop is minor and quality degradation can be hardly observed. Even with pruning ratios exceeding 90\%, our method consistently outperforms 3DGS without SHs by a considerable margin in terms of PSNR values. We also notice that 3DGS without SHs may blend colors from different viewpoints for specific objects, as evidenced by the black scratch artifacts appearing on the base of the silver ball. Our adaptive SH pruning successfully avoids these artifacts, even when the pruning ratio exceeds 90\%.

\subsubsection{Representation Composition.}

We dissect the compressed representation and analyze its composition. From scene Truck, we select 2 representations with different rates. The compressed representation consists of 5 components: header, indexes, codebooks, logits and positions, as listed in Tab.~\ref{tab:representation composition}. Specifically, the header consists of Gaussian indexes of SH mask cluster boundaries as mentioned in Sec.~\ref{subsubsec: rearranging gaussians} as well as the quantization step and the minimal opacity value used in opacity quantization. Logits represent the probability distribution of indexes which is needed for arithmetic coding.

It can be derived that positions and indexes dominate the proportion in the representation at high bitrates. As bitrate decreases, the proportion for codebook and logits gradually increases.

\begin{table}[tb]
  \caption{Composition of our high-rate Gaussian representation and low-rate one trained on scene Truck.}
  \label{tab:representation composition}
  \centering
  \begin{tabular}{@{}l@{\hspace{2mm}}|@{\hspace{2mm}}c@{\hspace{2mm}}|@{\hspace{2mm}}c@{\hspace{2mm}}|@{\hspace{2mm}}c@{\hspace{2mm}}|@{\hspace{2mm}}c@{}}
    \toprule
     &
    \multicolumn{2}{c@{\hspace{2mm}}|@{\hspace{2mm}}}{High Rate} &
    \multicolumn{2}{c}{Low Rate} \\
    \cmidrule{2-5}          Component   & Rate (MB)  & Proportion (\%)   & Rate (MB) & Proportion (\%)   \\
    \midrule
    Header      & $4\times10^{-5}$ & $\approx0\%$ & $4\times10^{-5}$ & $\approx0\%$ \\
    Indexes     & $7.23$   & $48.7\%$       & $0.54$      & $36.2\%$       \\
    Codebooks   & $0.98$   & $6.6\%$        & $0.32$      & $21.8\%$       \\
    Logits      & $0.12$   & $0.8\%$        & $0.09$      & $5.8\%$       \\
    Positions   & $6.53$   & $43.9\%$       & $0.53$      & $36.1\%$       \\
    \midrule
    Total       & $14.86$  & -              & $1.48$      & -            \\
  \bottomrule
  \end{tabular}
\end{table}

\section{Conclusion}
Our presented RDO-Gaussian formulates the learning of 3D Gaussian representations as a joint rate-distortion optimization process, achieving flexible and continuous rate control. Moreover, we devise an adaptive SHs pruning strategy that effectively removes redundant SHs while preserving view-dependent appearance. Experimental results indicate that our method can achieve over 40$\times$ compression ratio on 3DGS with minor distortion.

One of the limitations of our work is that representations are trained individually for different rates without shared information. This introduces additional storage overhead when the same scene is represented by multiple Gaussian representations of different bitrates. In future work, we aim to design a variable-rate Gaussian representation that supports multiple bitrates in a single representation.

\bibliographystyle{splncs04}
\bibliography{main}

% \newpage
\section*{Appendix}
\appendix

\setcounter{section}{0}
\setcounter{equation}{0}
\setcounter{figure}{0}
\setcounter{table}{0}

\renewcommand{\thesection}{A\arabic{section}}
\renewcommand{\thefigure}{A.\arabic{figure}}
\renewcommand{\thetable}{A.\arabic{table}}
\renewcommand{\theequation}{A.\arabic{equation}}

\newcommand{\fakeref}[1]{{\color{red}{#1}}}

In the appendix, we first provide a detailed description of our hyperparameters in Sec.~\ref{sec:hyperparameters}. Additionally, we present the quantitative results of our main experiments in Sec.~\ref{sec:quantitative results}. Furthermore, we conduct additional experiments to demonstrate the performance of our method in Sec.~\ref{sec:additional experiments}.

\section{Hyperparameters}
\label{sec:hyperparameters}
In this section, we present the hyperparameters utilized during training. For Gaussian pruning, we set the Gaussian mask threshold $\phi_{\rm thres}$ to $0.1$, and the learning rate of Gaussian masks to $0.01$. Regarding adaptive SHs pruning, the SH mask threshold $\theta_{\rm thres}$ is set to $0.1$, and the learning rate of SH masks for all $3$ degrees to $0.05$ for real scenes and $0.005$ for synthetic scenes. Concerning ECVQ, the learning rates are set to $0.0002$ for codebooks and $0.002$ for logits. The codebook size is $8192$ for scales, rotations and base colors, while it is $4096$ for SH parameters of all $3$ degrees. For $\lambda$ values across all modules, $\lambda_{\rm SHprune}$ and $\lambda_{\rm GSprune}$ differ in rates, as detailed in the first two rows of Tab.~\ref{tab:tandt quantitative comparison average} - Tab.~\ref{tab:ns quantitative comparison average}, whereas all $\lambda$ in ECVQ remain rate-invariant. Specifically, $\lambda^{(s)}$, which balances the rate and distortion for scales in ECVQ, is set to $32768$, while $\lambda^{(r)}$, $\lambda^{(DC)}$, $\lambda^{(SH1)}$, $\lambda^{(SH2)}$ and $\lambda^{(SH3)}$ are all set to $256$.

\section{Quantitative Results}
\label{sec:quantitative results}
In this section, we present the detailed experimental results for all four datasets. We organize the results into two groups. The first group displays the quantitative results averaged over scenes on each dataset across different rates, as listed in Tab.~\ref{tab:tandt quantitative comparison average} - Tab.~\ref{tab:ns quantitative comparison average}. The second group shows the quantitative results with the highest rate on each dataset across different scenes, as presented in Tab.~\ref{tab:tandtdb quantitative comparison per scene} - Tab.~\ref{tab:ns quantitative comparison per scene}.

\begin{table}[H]
  \caption{Quantitative results on Tanks and Temples dataset across different rates. The values are averaged over all scenes in the dataset.}
  \label{tab:tandt quantitative comparison average}
  \centering
  \begin{tabular}{@{}l@{\hspace{2mm}}|@{\hspace{2mm}}c@{\hspace{4mm}}c@{\hspace{4mm}}c@{\hspace{4mm}}c@{\hspace{4mm}}c@{\hspace{4mm}}c@{}}
    \toprule
    $\lambda_{\rm GSprune}$ & $0.05$ & $0.02$ & $0.01$ & $0.005$ & $0.002$  & $0.0005$ \\
    $\lambda_{\rm SHprune}$ & $0.5$  & $0.2$  & $0.1$  & $0.05$  & $0.02$   & $0.005$ \\
    \midrule
    Rate (MB)&1.32&2.39&3.74&5.49&8.02&12.02 \\
    PSNR&22.09&22.69&22.98&23.14&23.28&23.34 \\
    SSIM&0.755&0.793&0.812&0.823&0.831&0.835 \\
    LPIPS&0.318&0.264&0.233&0.214&0.202&0.195 \\
    Gaussian prune ratio&0.951&0.900&0.841&0.763&0.646&0.471 \\
    Ada. SHs prune ratio&0.998&0.993&0.985&0.970&0.930&0.833  \\
  \bottomrule
  \end{tabular}
\end{table}

\begin{table}[H]
  \caption{Quantitative results on Deep Blending dataset across different rates. The values are averaged over all scenes in the dataset.}
  \label{tab:db quantitative comparison average}
  \centering
  \begin{tabular}{@{}l@{\hspace{2mm}}|@{\hspace{2mm}}c@{\hspace{4mm}}c@{\hspace{4mm}}c@{\hspace{4mm}}c@{\hspace{4mm}}c@{\hspace{4mm}}c@{}}
    \toprule
    $\lambda_{\rm GSprune}$ & $0.05$ & $0.02$ & $0.01$ & $0.005$ & $0.002$  & $0.0005$ \\
    $\lambda_{\rm SHprune}$ & $0.5$  & $0.2$  & $0.1$  & $0.05$  & $0.02$   & $0.005$ \\
    \midrule
    Rate (MB)&1.22&2.40&4.14&6.71&10.96&18.00 \\
    PSNR&28.38&29.07&29.35&29.48&29.59&29.63 \\
    SSIM&0.872&0.887&0.895&0.899&0.901&0.902 \\
    LPIPS&0.331&0.296&0.277&0.265&0.256&0.252 \\
    Gaussian prune ratio&0.974&0.939&0.891&0.818&0.693&0.484 \\
    Ada. SHs prune ratio&0.999&0.998&0.995&0.987&0.965&0.894 \\
  \bottomrule
  \end{tabular}
\end{table}

\begin{table}[H]
  \caption{Quantitative results on Mip-NeRF 360 dataset across different rates. The values are averaged over all scenes in the dataset.}
  \label{tab:m360 quantitative comparison average}
  \centering
  \begin{tabular}{@{}l@{\hspace{2mm}}|@{\hspace{2mm}}c@{\hspace{4mm}}c@{\hspace{4mm}}c@{\hspace{4mm}}c@{\hspace{4mm}}c@{\hspace{4mm}}c@{}}
    \toprule
    $\lambda_{\rm GSprune}$ & $0.05$ & $0.02$ & $0.01$ & $0.005$ & $0.002$  & $0.0005$ \\
    $\lambda_{\rm SHprune}$ & $0.5$  & $0.2$  & $0.1$  & $0.05$  & $0.02$   & $0.005$ \\
    \midrule
    Rate (MB)&1.71&3.54&6.16&9.76&15.35&23.46 \\
    PSNR&24.43&25.38&26.03&26.50&26.87&27.05 \\
    SSIM&0.683&0.734&0.764&0.784&0.796&0.802 \\
    LPIPS&0.406&0.343&0.299&0.268&0.248&0.239 \\
    Gaussian prune ratio&0.965&0.922&0.862&0.776&0.636&0.437 \\
    Ada. SHs prune ratio&0.999&0.996&0.989&0.970&0.922&0.809 \\
  \bottomrule
  \end{tabular}
\end{table}

\begin{table}[H]
  \caption{Quantitative results on NeRF Synthetic dataset across different rates. The values are averaged over all scenes in the dataset.}
  \label{tab:ns quantitative comparison average}
  \centering
  \begin{tabular}{@{}l@{\hspace{2mm}}|@{\hspace{2mm}}c@{\hspace{4mm}}c@{\hspace{4mm}}c@{\hspace{4mm}}c@{\hspace{4mm}}c@{\hspace{4mm}}c@{}}
    \toprule
    $\lambda_{\rm GSprune}$ & $0.005$ & $0.002$ & $0.001$ & $0.0005$ & $0.0002$ & $0.0001$  \\
    $\lambda_{\rm SHprune}$ & $0.025$ & $0.01$  & $0.005$ & $0.0025$ & $0.001$  & $0.0005$  \\
    \midrule
    Rate (MB)&0.42&0.73&1.02&1.33&1.84&2.31 \\
    PSNR&29.95&31.23&32.01&32.56&32.97&33.12 \\
    SSIM&0.946&0.956&0.961&0.964&0.966&0.967 \\
    LPIPS&0.064&0.050&0.043&0.038&0.035&0.035 \\
    Gaussian prune ratio&0.942&0.895&0.840&0.768&0.644&0.532 \\
    Ada. SHs prune ratio&0.961&0.864&0.749&0.624&0.473&0.374 \\
  \bottomrule
  \end{tabular}
\end{table}

\begin{table}[H]
  \caption{Quantitative comparison on Tanks and Temples dataset and Deep Blending dataset across different scenes. The results of our method are from the highest rate configuration. 3DGS is implemented by ourselves.}
  \label{tab:tandtdb quantitative comparison per scene}
  \centering
  \begin{tabular}{@{}l@{\hspace{2mm}}|@{\hspace{2mm}}l@{\hspace{2mm}}|@{\hspace{2mm}}c@{\hspace{2mm}}c@{\hspace{2mm}}c@{\hspace{2mm}}c@{}}
    \toprule
    &Scene&Truck&Train&DrJohnson&Playroom \\
    \midrule
    \multirow{4}{*}{3DGS} &Rate (MB)&609.32&254.89&771.44&550.56 \\
    &PSNR&25.36&21.97&29.06&29.96 \\
    &SSIM&0.878&0.810&0.898&0.901 \\
    &LPIPS&0.148&0.209&0.247&0.246 \\
    \midrule
    \multirow{4}{*}{Ours} &Rate (MB)&14.87&9.18&22.89&13.11 \\
    &PSNR&25.04&21.64&29.10&30.16 \\
    &SSIM&0.870&0.799&0.899&0.905 \\
    &LPIPS&0.161&0.228&0.253&0.251 \\
  \bottomrule
  \end{tabular}
\end{table}

\vspace{-5mm}

\begin{table}[H]
  \caption{Quantitative comparison on Mip-NeRF 360 dataset (ourdoor scenes) across different scenes. The results of our method are from the highest rate configuration. 3DGS is implemented by ourselves.}
  \label{tab:m360 ourdoor quantitative comparison per scene}
  \centering
  \begin{tabular}{@{}l@{\hspace{2mm}}|@{\hspace{2mm}}l@{\hspace{2mm}}|@{\hspace{2mm}}c@{\hspace{2mm}}c@{\hspace{2mm}}c@{\hspace{2mm}}c@{\hspace{2mm}}c@{}}
    \toprule
    &Scene&Bicycle&Flowers&Garden&Stump&Treehill \\
    \midrule
    \multirow{4}{*}{3DGS} &Rate (MB)&1335.48&807.02&1329.37&1096.31&809.95 \\
    &PSNR&25.11&21.32&27.30&26.67&22.51 \\
    &SSIM&0.747&0.589&0.856&0.770&0.635 \\
    &LPIPS&0.244&0.359&0.122&0.242&0.348 \\
    \midrule
    \multirow{4}{*}{Ours} &
    Rate (MB)&43.14&26.52&38.75&36.17&28.76 \\
    &PSNR&24.88&21.16&26.78&26.53&22.53 \\
    &SSIM&0.734&0.577&0.838&0.762&0.629 \\
    &LPIPS&0.266&0.372&0.148&0.258&0.364 \\
  \bottomrule
  \end{tabular}
\end{table}

\vspace{-5mm}

\begin{table}[H]
  \caption{Quantitative comparison on Mip-NeRF 360 dataset (indoor scenes) across different scenes. The results of our method are from the highest rate configuration. 3DGS is implemented by ourselves.}
  \label{tab:m360 indoor quantitative comparison per scene}
  \centering
  \begin{tabular}{@{}l@{\hspace{2mm}}|@{\hspace{2mm}}l@{\hspace{2mm}}|@{\hspace{2mm}}c@{\hspace{2mm}}c@{\hspace{2mm}}c@{\hspace{2mm}}c@{}}
    \toprule
    &Scene&Room&Counter&Kitchen&Bonsai \\
    \midrule
    \multirow{4}{*}{3DGS} &Rate (MB)&349.48&275.71&415.42&293.48 \\
    &PSNR&31.62&29.10&31.45&32.33 \\
    &SSIM&0.926&0.914&0.932&0.946 \\
    &LPIPS&0.197&0.184&0.117&0.181 \\
    \midrule
    \multirow{4}{*}{Ours} &
    Rate (MB)&8.11&7.91&13.01&8.77 \\
    &PSNR&31.11&28.51&30.49&31.43 \\
    &SSIM&0.918&0.902&0.921&0.937 \\
    &LPIPS&0.212&0.205&0.130&0.196 \\
  \bottomrule
  \end{tabular}
\end{table}

\begin{table}[H]
  \caption{Quantitative comparison on NeRF Synthetic dataset across different scenes. The results of our method are from the highest rate configuration. 3DGS is implemented by ourselves.}
  \label{tab:ns quantitative comparison per scene}
  \centering
  \begin{tabular}{@{}l@{\hspace{2mm}}|@{\hspace{2mm}}l@{\hspace{2mm}}|@{\hspace{2mm}}c@{\hspace{2mm}}c@{\hspace{2mm}}c@{\hspace{2mm}}c@{\hspace{2mm}}c@{\hspace{2mm}}c@{\hspace{2mm}}c@{\hspace{2mm}}c@{}}
    \toprule
    &Scene&Chair&Drums&Ficus&Hotdog&Lego&Mic&Materials&Ship \\
    \midrule
    \multirow{4}{*}{3DGS} & Rate (MB)&115.68&92.29&62.40&43.92&81.41&45.87&37.74&65.76 \\
    & PSNR&35.55&26.28&35.48&38.03&36.10&36.70&30.48&31.67 \\
    & SSIM&0.988&0.955&0.987&0.985&0.983&0.993&0.960&0.906 \\
    & LPIPS&0.010&0.037&0.012&0.020&0.016&0.006&0.037&0.106 \\
    \midrule
    \multirow{4}{*}{Ours} & Rate (MB)&2.98&3.02&2.05&1.66&2.77&1.46&1.91&2.65 \\
    & PSNR&34.63&26.01&35.21&37.28&35.25&35.60&29.94&31.07 \\
    & SSIM&0.984&0.953&0.986&0.983&0.980&0.991&0.957&0.901 \\
    & LPIPS&0.014&0.040&0.013&0.024&0.020&0.008&0.043&0.115 \\
  \bottomrule
  \end{tabular}
\end{table}

\section{Additional Experiments}
\label{sec:additional experiments}
\subsection{Effect of Gaussian Pruning}

\begin{figure}[h]
    \centering
    \begin{subfigure}{1\linewidth}
        \includegraphics[width=1\textwidth]{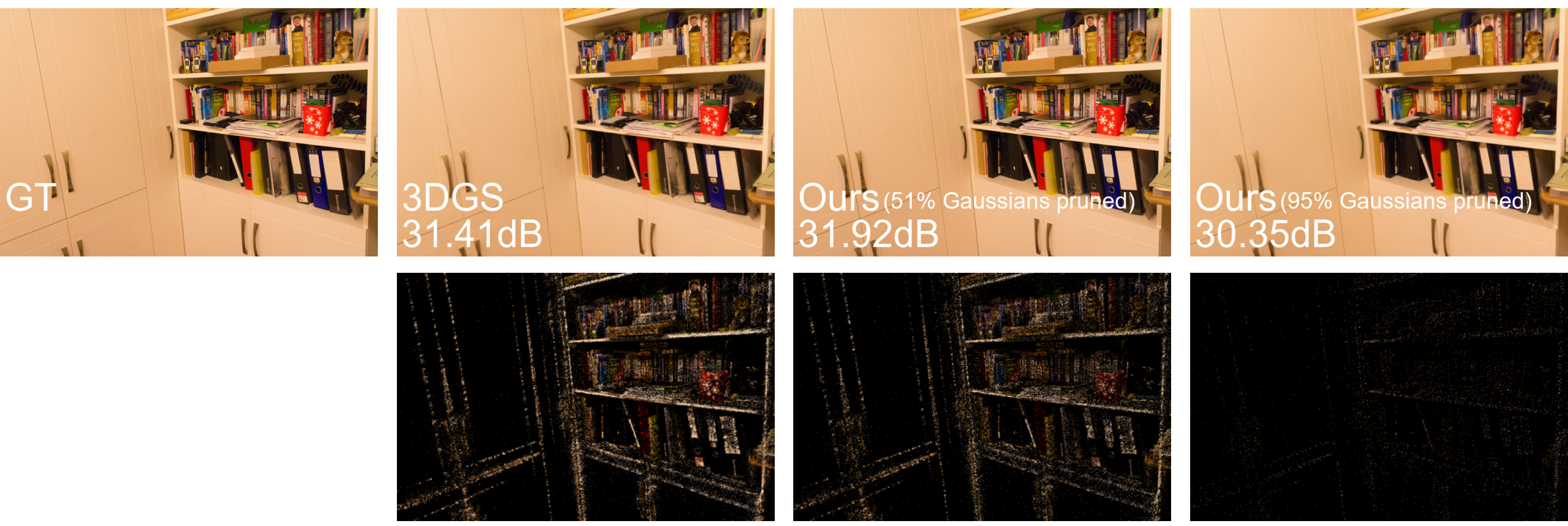}
    \end{subfigure}
    \begin{subfigure}{1\linewidth}
        \includegraphics[width=1\textwidth]{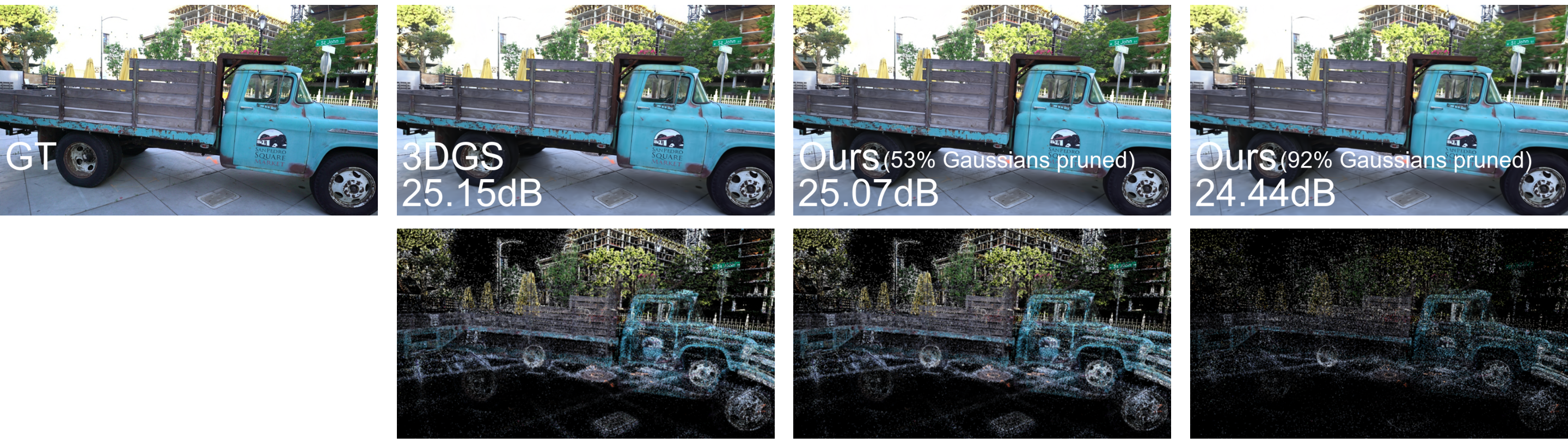}
    \end{subfigure}
    \caption{Visualization of Gaussian pruning. We show the rendered images as well as the point clouds of our method with two different Gaussian prune ratio and 3DGS on scene Playroom and Truck. Each point in the point cloud represents a scaled Gaussian ellipsoid.}
    \label{fig:gsprune}
\end{figure}

To validate the performance of Gaussian pruning, we deactivate other modules and exclusively enable Gaussian pruning. We vary $\lambda_{\rm GSprune}$ values for different prune ratios. The results are depicted in Fig.~\ref{fig:gsprune}. We observe that our method can attain the quality of 3DGS when approximately $50\%$ Gaussians are pruned. Moreover, even with a prune ratio exceeding $90\%$, our method can still depict the scene using sparser Gaussians without significant quality degradation.

\subsection{Rendering Speed}

We compare the average Frames Per Second (FPS) of our method and 3DGS during the rendering process on real-scene datasets, as outlined in Tab.~\ref{tab:fps}. Both methods are implemented on an NVIDIA A100 GPU. As our compressed representation contains fewer Gaussian points than 3DGS, a higher FPS can be achieved. Furthermore, as the rate decreases, resulting in fewer Gaussians, the FPS increases accordingly.

\begin{table}[ht]
  \caption{FPS of our method with different rates and 3DGS on real-scene datasets. Our methods are tested with different rate configurations (denoted as Rate 1-6, from high rate to low rate).}
  \label{tab:fps}
  \centering
  \begin{tabular}{@{}l@{\hspace{2mm}}|@{\hspace{2mm}}l@{\hspace{2mm}}|@{\hspace{2mm}}c@{\hspace{2mm}}c@{\hspace{2mm}}c@{}}
    \toprule
    \multicolumn{2}{c@{\hspace{2mm}}|@{\hspace{2mm}}}{Dataset}&Mip-NeRF 360&Tanks and Temples&Deep Blending \\
    \midrule
    \multicolumn{2}{c@{\hspace{2mm}}|@{\hspace{2mm}}}{3DGS} & 141 & 176 & 143 \\
    \midrule
    \multirow{6}{*}{Ours} & Rate 1 & 191 & 269 & 207 \\
    &Rate 2 & 244 & 367 & 302 \\
    &Rate 3 & 308 & 444 & 395 \\
    &Rate 4 & 361 & 531 & 485 \\
    &Rate 5 & 445 & 636 & 572 \\
    &Rate 6 & 509 & 732 & 710 \\
  \bottomrule
  \end{tabular}
\end{table}

\subsection{Index Composition}
We analyze the composition of the indexes in our compressed representation. As detailed in Tab.~\ref{tab:index composition}, the indexes comprise 7 attributes: scales, rotations, base colors, SHs of degree 1-3 and opacities. Among these attributes, scales, rotations, and base colors predominantly contribute to the proportion of the indexes. As the bitrate decreases, the proportion of SHs decreases due to the higher prune ratio applied to SHs at lower bitrates.

\begin{table}[ht]
  \caption{Index composition of our high-rate Gaussian representation and low-rate one trained on scene Truck.}
  \label{tab:index composition}
  \centering
  \begin{tabular}{@{}l@{\hspace{2mm}}|@{\hspace{2mm}}c@{\hspace{2mm}}@{\hspace{2mm}}c@{\hspace{2mm}}|@{\hspace{2mm}}c@{\hspace{2mm}}@{\hspace{2mm}}c@{}}
    \toprule
     &
    \multicolumn{2}{c@{\hspace{2mm}}|@{\hspace{2mm}}}{High Rate} &
    \multicolumn{2}{c}{Low Rate} \\
    \cline{2-5}         Attribute   & Rate (MB)  & Proportion (\%)   & Rate (MB) & Proportion (\%)   \\
    \midrule
    Scales            & $1.83$   & $25.27\%$  & $0.151$            & $28.16\%$ \\
    Rotations         & $1.89$   & $26.19\%$ & $0.147$             & $27.42\%$       \\
    Base colors       & $1.84$   & $25.48\%$ & $0.148$             & $27.60\%$       \\
    SHs of degree 1   & $0.14$   & $2.06\%$  & $8.0\times10^{-5}$  & $0.01\%$       \\
    SHs of degree 2   & $0.20$   & $2.86\%$  & $1.00\times10^{-4}$ & $0.02\%$       \\
    SHs of degree 3   & $0.26$   & $3.65\%$  & $1.76\times10^{-4}$ & $0.03\%$\\
    Opacities         & $1.04$   & $14.50\%$ & $0.089$             & $16.76\%$\\
    \midrule
    Total             & $7.23$  & -          & $0.537$      & -            \\
  \bottomrule
  \end{tabular}
\end{table}

\end{document}